\documentclass[conference]{IEEEtran}
\usepackage{cite}
\usepackage{amsmath,amssymb,amsfonts}
\usepackage{algorithmic}
\usepackage{graphicx}
\usepackage[countmax]{subfloat}
\usepackage[caption=false,font=footnotesize]{subfig} % Correct setup for using subfig
\usepackage{tabularx}
\usepackage{url}
\usepackage{textcomp}
\usepackage{todonotes}
\usepackage{color}
\usepackage{xcolor}
\usepackage{soul}
\usepackage[most]{tcolorbox}
\usetikzlibrary{shapes.geometric, arrows.meta, positioning}

\newtcbox{\tightboxy}[1][]{on line,
  colframe=white, % Frame color
  colback=yellow, % Background color, change as needed
  boxrule=0pt, % No border
  arc=0pt, % No rounded corners
  boxsep=0pt, % No separation between text and box border
  left=0pt, % Padding inside the box on the left (adjust as needed)
  right=0pt, % Padding inside the box on the right
  top=2pt, % Padding inside the box on the top
  bottom=2pt, % Padding inside the box on the bottom
  tcbox raise base
}

\newtcbox{\tightboxg}[1][]{on line,
  colframe=white, % Frame color
  colback=green, % Background color, change as needed
  boxrule=0pt, % No border
  arc=0pt, % No rounded corners
  boxsep=0pt, % No separation between text and box border
  left=0pt, % Padding inside the box on the left (adjust as needed)
  right=0pt, % Padding inside the box on the right
  top=2pt, % Padding inside the box on the top
  bottom=2pt, % Padding inside the box on the bottom
  tcbox raise base
}

\newtcbox{\tightboxr}[1][]{on line,
  colframe=white, % Frame color
  colback=red, % Background color, change as needed
  boxrule=0pt, % No border
  arc=0pt, % No rounded corners
  boxsep=0pt, % No separation between text and box border
  left=0pt, % Padding inside the box on the left (adjust as needed)
  right=0pt, % Padding inside the box on the right
  top=2pt, % Padding inside the box on the top
  bottom=2pt, % Padding inside the box on the bottom
  tcbox raise base
}

\def\BibTeX{{\rm B\kern-.05em{\sc i\kern-.025em b}\kern-.08em
   T\kern-.1667em\lower.7ex\hbox{E}\kern-.125emX}}

\begin{document}

\title{
Generative AI for Research Data Processing: Lessons Learnt From Three Use Cases}
%{\footnotesize \textsuperscript{*}Note: Sub-titles are not captured in Xplore and should not be used}

\author{\IEEEauthorblockN{1\textsuperscript{st} Modhurita Mitra}
\IEEEauthorblockA{
\textit{Utrecht University}\\
Utrecht, The Netherlands \\
m.mitra@uu.nl}
\and
\IEEEauthorblockN{2\textsuperscript{nd} Martine G. de Vos}
\IEEEauthorblockA{
\textit{Utrecht University}\\
Utrecht, The Netherlands \\
m.g.devos@uu.nl}
\and
\IEEEauthorblockN{3\textsuperscript{rd} Nicola Cortinovis}
\IEEEauthorblockA{
\textit{Utrecht University}\\
Utrecht, The Netherlands \\
n.cortinovis@uu.nl}
\and
\IEEEauthorblockN{4\textsuperscript{th} Dawa Ometto}
\IEEEauthorblockA{
\textit{Utrecht University}\\
Utrecht, The Netherlands \\
d.l.a.ometto1@uu.nl}
}

\maketitle

\begin{abstract}
There has been enormous interest in generative AI since ChatGPT was launched in 2022. However, there are concerns about the accuracy and consistency of the outputs of generative AI. We have carried out an exploratory study on the application of this new technology in research data processing. We identified tasks for which rule-based or traditional machine learning approaches were difficult to apply, and then performed these tasks using generative AI. 

We demonstrate the feasibility of using the generative AI model Claude 3 Opus in three research projects involving complex data processing tasks: 
\begin{enumerate}
\item \textit{Information extraction}: We extract plant species names from historical seedlists (catalogues of seeds) published by botanical gardens. 
\item \textit{Natural language understanding}: We extract certain data points (name of drug, name of health indication, relative effectiveness, cost-effectiveness, etc.) from documents published by Health Technology Assessment organisations in the EU. 
\item \textit{Text classification}: We assign industry codes to projects on the crowdfunding website Kickstarter. 
\end{enumerate}

We share the lessons we learnt from these use cases: How to determine if generative AI is an appropriate tool for a given data processing task, and if so, how to maximise the accuracy and consistency of the results obtained.

%This document is a model and instructions for \LaTeX.
%This and the IEEEtran.cls file define the components of your paper [title, text, heads, etc.]. *CRITICAL: Do Not Use Symbols, Special Characters, Footnotes, 
%or Math in Paper Title or Abstract.
\end{abstract}


\section{Introduction}
In this paper, we share our insights on the application of generative AI in research software engineering projects. 

Generative AI can potentially be used to perform a wide variety of research data processing tasks, such as interpreting documents, extracting information from them, and classifying text into categories. Since the tasks are specified through prompts in natural language, the barrier to entry is low. Therefore, this tool can be easily used by domain experts in a wide range of fields, with varying levels of programming skills and depth of knowledge of technical topics such as machine learning. 

As a research software engineering team, we have seen a growing interest among researchers at our university in applying generative AI in their projects. Recent publications, policy documents, and legislation
address the opportunities and risks of generative AI systems and provide guidelines on how to deal with this new technology (e.g. \cite{Ferrari2023, EU2024guidelines, EU2024AIact}). 
However, the broad and general nature of these recommendations means that they do not provide much concrete guidance. 

Generative AI tools such as ChatGPT\footnote{\url{https://chatgpt.com/}} have until now been used in scientific contexts mainly in ``supervised mode", for tasks such as editing manuscripts, making presentation slides, brainstorming ideas, assisting in literature reviews, and helping to write or check code \cite{Stokel-Walker2023, VanDis2023, Dwivedi2023}.
A human reviews the output of generative AI and decides if
it is acceptable, and possibly issues further prompts to obtain
more refined results from generative AI. When dealing with
large amounts of data, this kind of iterative human supervision of
every single data sample is not possible. We focus on the use
of generative AI for performing complex operations, such as
parsing, transforming, and extracting information from large
amounts of textual data, in ``unsupervised mode", without a ``human-in-the-loop" \cite{Nah2023, Mosqueira-Rey2023}.

In this study, we explore the extent to which the application of generative AI affects the execution and results of research projects.
Generative AI can produce erroneous and inconsistent outputs, due to its generative variability and/or hallucination \cite{lee2024one, hallucination}. Therefore, we pay particular attention to the accuracy and consistency of the results.
We formulate concrete considerations that can be used to determine whether generative AI can be applied in a particular research project.
To this end we apply generative AI in three research projects and for each evaluate both the engineering process and the results produced.

\subsection{Goal}
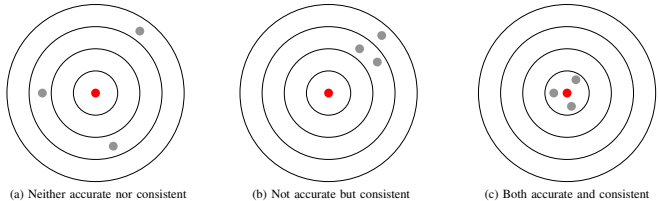
\begin{figure}[htbp]
    \centering
    \resizebox{0.48\textwidth}{!}{%
    \subfloat[Neither accurate nor consistent]{%
        \begin{tikzpicture}
            \draw (0, 0) circle (0.5);
            \draw (0, 0) circle (1);
            \draw (0, 0) circle (1.5);
            \draw (0, 0) circle (2);
            \fill[red] (0, 0) circle (0.1);
            \fill[gray,opacity=0.85] (-1.2, 0) circle (0.1);
            \fill[gray,opacity=0.85] (1.0, 1.4) circle (0.1);
            \fill[gray,opacity=0.85] (0.4, -1.2) circle (0.1);
        \end{tikzpicture}
        \label{fig:not_accurate_not_consistent}
    }
    \hspace{1cm}
    \subfloat[Not accurate but consistent]{%
        \begin{tikzpicture}
            \draw (0, 0) circle (0.5);
            \draw (0, 0) circle (1);
            \draw (0, 0) circle (1.5);
            \draw (0, 0) circle (2);
            \fill[red] (0, 0) circle (0.1);
            \fill[gray,opacity=0.85] (0.7, 1) circle (0.1);
            \fill[gray,opacity=0.85] (1.2, 1.3) circle (0.1);
            \fill[gray,opacity=0.85] (1.1, 0.7) circle (0.1);
        \end{tikzpicture}
        \label{fig:consistent_but_not_accurate}
    }
    \hspace{1cm}
    \subfloat[Both accurate and consistent]{
        \begin{tikzpicture}
            \draw (0, 0) circle (0.5);
            \draw (0, 0) circle (1);
            \draw (0, 0) circle (1.5);
            \draw (0, 0) circle (2);
            \fill[red] (0, 0) circle (0.1);
            \fill[gray,opacity=0.85] (-0.3, 0) circle (0.1);
            \fill[gray,opacity=0.85] (0.2, 0.3) circle (0.1);
            \fill[gray,opacity=0.85] (0.1, -0.3) circle (0.1);
        \end{tikzpicture}
        \label{fig:accurate_and_consistent}
    }
    }
    \caption{Accuracy and consistency of outputs: The red dot in the centre is the true value or the ground truth, and the three grey dots indicate outputs produced by different generative AI runs with the same input data, model, and parameters. For generative AI to be a reliable data processing method, we want our results to look like Fig. \ref{fig:accurate_and_consistent}, in which the outputs are both accurate and consistent.}
    \label{fig:accuracy_consistency}
\end{figure}

The goal of this study is twofold: 
\begin{enumerate}
\item To determine the conditions under which generative AI is an appropriate tool for a given research task, and 
\item To determine strategies to maximise the accuracy and consistency of the results obtained using generative AI.
\end{enumerate}
We focus on two aspects that are both crucial to research data processing methods \cite{data_consistency_accuracy, software_accuracy_reliability, reproducibility2019}, and about which concerns have been raised with regard to generative AI (Table \ref{tab:lit-concerns}): \textit{accuracy} and \textit{consistency} of the results obtained using this technique. We define accuracy of the results as the closeness of the agreement between the results from generative AI with the ``true result", or a so-called ``ground truth". We define consistency of the results as the closeness of the agreement of the results with each other when the same task is performed multiple times with the same data, using the same generative AI model and model parameters. For generative AI to be a \textit{reliable} data processing tool, the results must be both accurate and consistent. The concepts of accuracy and consistency are illustrated in Fig. \ref{fig:accuracy_consistency}. 

At this exploratory stage, we present our findings qualitatively, via illustrative examples, and not via traditional machine learning performance metrics such as precision and recall. We believe that such examples serve to illustrate the power and pitfalls of generative AI better than numerical values of metrics.

\subsection{Structure of the paper}
\label{subsec:structure}

In Section \ref{sec:genai_concerns} we present an introduction to generative AI, concerns about its use, and a literature survey of its use in research. In Section \ref{sec:method} we explain how we use generative AI in this study. In Sections \ref{sec:seedlists}, \ref{sec:hta}, and \ref{sec:kickstarter} we present the three use cases. For each of these use cases we describe the scientific goal of the project, the research engineering task to be performed, and the input data. We then argue why generative AI is a good choice for performing this research engineering task, and how we perform this task using generative AI. We then present the results and evaluate them by comparing the accuracy and consistency of the results against a manually-constructed ground truth. In Section \ref{sec:discussion} we discuss the common themes connecting the three use cases and the general principles we have arrived at about the use of generative AI for research data processing tasks. We then 
briefly describe our plans for future work.

\section{Generative AI, potential concerns, and related work}
\label{sec:genai_concerns}
% Generative AI
Generative AI is AI that can produce text, image, audio, or video output in response to a user-provided input prompt. Generative AI models such as GPT and Claude are based on Large Language Models (LLMs), which are deep neural networks based on the transformer architecture \cite{NIPS2017_3f5ee243}. 
These models are trained on much of the publicly-available data on the internet, non-public data from other sources, and their own internally-generated data \cite{gpt4systemcard, claude3modelcard}. These are then further fine-tuned through reinforcement learning from human feedback (RLHF) \cite{NEURIPS2022_b1efde53_short}. 

Generative AI is a general-purpose technology that can potentially be used in a wide variety of fields \cite{eloundou2023gpts}. However, it needs to be appropriately customised for the task at hand. A social media content creator might want to use generative AI to create content that is varied and creative. On the other hand, for analytical tasks in research data processing, the creativity and variability of generative AI need to be minimised to maximise the accuracy and consistency of the results.

% Concerns
There are serious and legitimate concerns about generative AI that have led to efforts to govern AI, especially in the European Union (EU) \cite{EU2024AIact, pmid38366218, EU2024guidelines}. Many concerns relate to the way generative AI models are trained, documented, and published. There is discomfort in the research community about the lack of transparency with regard to both the data and the deep learning architecture, lack of reproducibility of outputs, encoded bias, and sustainability issues (see Table \ref{tab:lit-concerns} for a list of concerns).

While we consider these concerns important, we focus on the \textit{generative variability} of generative AI, which distinguishes it from traditional discriminative AI systems -- \textit{generative} because the goal of generative AI applications is to produce artefacts as outputs, rather than decision boundaries and \textit{variability} because for a given input these systems can produce a variety of possible outputs \cite{Weisz2023a,Megahed2023}. Also, generative AI may produce text that can be convincing but inaccurate \cite{hallucination}, so its use can distort scientific facts and spread misinformation \cite{VanDis2023}. This can happen with specialised topics that the model might have had little data to train on \cite{Stokel-Walker2023,Pal2023}. The systematic and structured construction of prompts may help to control the output \cite{Weisz2023a, AlKhamissi2022}. However, the inconsistency is part of these systems since the foundation models, Pretrained Language Models, already show poor consistency in representing factual knowledge \cite{Elazar2021}. 

\textit{Computational reproducibility} is generally desired and expected of scientific studies using digital data \cite{OpenResearch2015_short}. A different researcher should be able to obtain the same results using the same input data, models, and code as the ones used in the original study \cite{reproducibility2019}. Proprietary model providers like OpenAI routinely deprecate older models in favour of new ones. This means that the generative AI models and the corresponding public APIs used for a particular study might no longer be available at a later time, and thus the results can no longer be reproduced. This can be avoided by using open-weights models like the Llama series of foundation models from Meta AI \cite{meta_llama}. These models can be downloaded and run locally, but this requires significant computational resources and technical know-how, which raises the barrier to entry for domain researchers' use of generative AI. However, this might be the only feasible option when processing private or sensitive data, which one might not want to send across the internet to a proprietary model provider while using their API.

A study in finance research used ChatGPT for assistance during multiple stages of the research process: idea generation, literature review, data identification and processing, and selection and implementation of a suitable testing framework \cite{Dowling2023}. They find that while the results produced by ChatGPT were plausible, addition of private data and domain expertise greatly improved the quality of the results.

A study in Statistical Process Control found that while ChatGPT performed well for well-defined tasks such as converting code from one programming language to another and providing explanations for simple concepts, it failed to satisfactorily perform difficult tasks such as writing new code from scratch and explaining lesser-known concepts \cite{Megahed2023}. They conclude that the output from ChatGPT needs to be validated for accuracy by human experts.

During our literature search we did not find any studies on the systematic application of generative AI for research data processing. This makes our study novel; we fill this gap by applying generative AI to three use cases from different research domains and deriving insights from this process.

\begin{table}
\centering
\caption{Main concerns regarding generative AI}
\label{tab:lit-concerns}
\begin{tabular}{|l|l|}
    \hline
    \textbf{Concern} & \textbf{References}\\
    \hline
    Lack of transparency & \cite{Bender2021,VanDis2023,Deelman2023,Weisz2023a,Dwivedi2023} \\
    Encoded bias & \cite{Bender2021,VanDis2023,Deelman2023,Weisz2023a,Dwivedi2023,Stokel-Walker2023,Pal2023} \\    
    No acknowledgement of sources & \cite{VanDis2023,Weisz2023a,Dwivedi2023,Stokel-Walker2023} \\ 
    Erroneous output &  \cite{VanDis2023,Weisz2023a,Dwivedi2023,Stokel-Walker2023,Pal2023,Megahed2023,hallucination} \\ 
    Inconsistent output & \cite{VanDis2023,Weisz2023a,Elazar2021,Megahed2023} \\   
    Environmental costs & \cite{Bender2021,Stokel-Walker2023,Deelman2023,Luccioni2024}\\   
    \hline
\end{tabular}
\end{table}

\section{Method}
\label{sec:method}

\begin{figure}[htbp]
\centering
\begin{tikzpicture}[auto, node distance=0.6cm, >=Latex, scale=0.8, transform shape]
    % Define styles
    \tikzset{
        startstop/.style={rectangle, rounded corners, minimum width=3cm, minimum height=1cm, align=center, draw=black},
        inout/.style={trapezium, trapezium left angle=70, trapezium right angle=110, minimum width=3cm, minimum height=1cm, align=center, draw=black},
        process/.style={rectangle, minimum width=3cm, minimum height=1cm, align=center, draw=black},
        decision/.style={diamond, aspect=2, minimum width=3cm, minimum height=1cm, align=center, draw=black},
        arrow/.style={thick,->},
        tight/.style={node distance=1.2cm} 
    }

    % Nodes
    \node (in1) [inout] {Input data};
    \node (pro1) [process, below=of in1] {Preprocessing: \\ Chunk data to satisfy\\ context length constraints};
    \node (sel1) [process, below=of pro1] {Select a chunk};
    \node (pro2) [process, below=of sel1] {Add instructions to create prompt};
    \node (pro3) [process, below=of pro2] {Process data with generative AI};
    \node (dec1) [decision, below=of pro3, node distance=1cm] {Error handling \\and retrying needed?}; 
    \node (pro4) [process, below=of dec1, node distance=1cm] {Post-processing}; 
    \node (dec2) [decision, below=of pro4, node distance=1cm] {More chunks to process?}; 
    \node (out1) [inout, below=of dec2] {Output data};

    % Arrows
    \draw [arrow] (in1) -- (pro1);
    \draw [arrow] (pro1) -- (sel1);
    \draw [arrow] (sel1) -- (pro2);
    \draw [arrow] (pro2) -- (pro3);
    \draw [arrow] (pro3) -- (dec1);
    \draw [arrow] (dec1) --  node[right] {No} (pro4);
    \draw [arrow] (dec1.east) -- node[above] {Yes} ++(1.2,0) |- (pro3.east); 
    \draw [arrow] (pro4) -- (dec2);
    \draw [arrow] (dec2.west) -- node[above] {Yes} ++(-1.2,0) |- (sel1.west); 
    \draw [arrow] (dec2) -- node[right] {No}  (out1);

\end{tikzpicture}
\caption{Generative AI pipeline}
\label{fig:pipeline}
\end{figure}
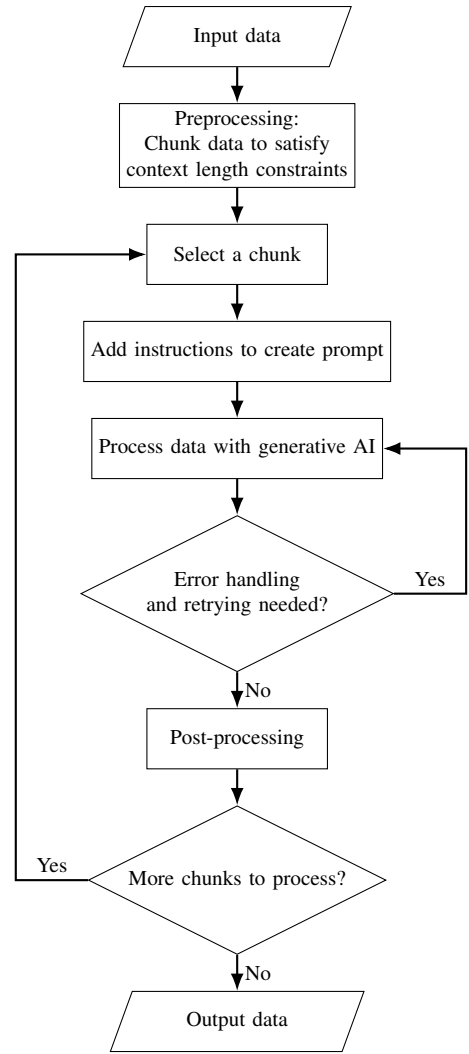

We apply generative AI in three research projects, listed and summarised in Table \ref{tab:use-cases}. Fig. \ref{fig:pipeline} shows our data processing pipeline. We first preprocess the input data by chunking it so that the number of tokens in the prompt is lower than the maximum number of input tokens allowed by the generative AI model. At this point we also estimate the size of the output and make sure that the expected output will contain fewer tokens than the maximum number of output tokens allowed. We then select one data chunk, and add instructions for data processing to create the prompt. We then supply this prompt to generative AI, which performs the data processing for this chunk. We then check if the data processing step was properly executed. If it was not properly executed, for example because of network timeout issues or exceeding the per-minute token limit, we need to perform error handling and retrying -- we repeat the data processing step with generative AI. Otherwise we move on to the post-processing step. At this step we extract the data we need from the output returned by generative AI. This step is needed because the output is sometimes not in an immediately machine-readable format -- there might be extraneous text such as ``Here is the output you requested", or other such text or explanation that we do not want as part of the structured output we are seeking. Thereafter we check if there are more chunks to process -- if there are more chunks left, we process the next chunk. When all the chunks are processed, we have our final output data.  

\begin{table*}[htbp]
\centering
\caption{Use cases overview}
\label{tab:use-cases}
    \centering
\begin{tabularx}{\textwidth}{|
  >{\raggedright\arraybackslash\hsize=.5\hsize}X | 
  >{\raggedright\arraybackslash\hsize=\hsize}X | 
  >{\raggedright\arraybackslash\hsize=1.5\hsize}X | 
  >{\raggedright\arraybackslash\hsize=.5\hsize}X | 
  >{\raggedright\arraybackslash\hsize=.5\hsize}X |
    >{\raggedright\arraybackslash\hsize=\hsize}X |
}
    \hline
    \textbf{Project} & \textbf{Scientific discipline / Academic department} & \textbf{Task} & \textbf{Type of task} & \textbf{Input datafile format} & \textbf{Amount of data to be processed} \\
    \hline
    Seedlists & Botanical Gardens & Extract plant species names from historical seedlists (catalogues of seeds) published by botanical gardens & Information extraction & PDF, Excel, OCR text from scanned documents & $\sim$30,000 seedlists \\
    & & & & & \\
    Health Technology Assessment (HTA) documents & Pharmaceutical Sciences & Extract certain data points from documents published by HTA organisations in the EU & Natural language understanding & PDF & A few thousand HTA documents\\
    & & & & & \\
    Kickstarter & Economic Geography & Assign industry codes to projects on the crowdfunding website Kickstarter & Text classification & CSV & $\sim$300,000 Kickstarter projects \\
    \hline
\end{tabularx}
\end{table*}

\subsection*{Configuration}
The following general considerations apply to all three projects:

\begin{enumerate}

\item \textit{Temperature:} Temperature is a generative AI model parameter that can be set to control the randomness and variability of the outputs \cite{Anthropic_Glossary}. Lower values of temperature result in more conservative and predictable outputs, while higher values result in more creative and diverse outputs. The value of the temperature parameter is between 0 and 2. In this study, the results we seek are objective and well-defined. Therefore, we always set the temperature parameter to 0 to minimise randomness and variability of the outputs. This maximises both accuracy and consistency of our results. This choice aims to steer the tradeoff between exploration (of new possibilities) and exploitation (of existing knowledge) \cite{March1991-rp} towards the exploitation end of the spectrum. It should be emphasised that setting the temperature to 0 minimises non-determinism of the outputs, but does not completely eliminate it.

\item \textit{Machine-readable output:} We want the output to be machine-readable so that it can be automatically processed by a computer. Therefore, we ask generative AI to produce the output in JSON format.\footnote{https://www.json.org/json-en.html}

\item \textit{Generative AI models:} We used the Claude 3 Opus model from Anthropic AI. We used this model because we found by trial-and-error that it produced the most accurate and consistent results among the models we tried. The other models we tried were the GPT-3.5 and GPT-4 series of models from OpenAI. We accessed these models through these foundation model providers' public APIs.\footnote{\url{https://www.anthropic.com/api}}\textsuperscript{,}\footnote{\url{https://platform.openai.com/docs/overview}}

\end{enumerate}

The data, prompts, codes, and results for these use cases are available on GitHub.\footnote{https://github.com/UtrechtUniversity/generative-ai} The configuration settings  and the locations of the results for the examples reported in this paper are summarised in Table \ref{tab:settings}.

\begin{table*}[htbp]
\centering
\caption{Configuration settings}
\label{tab:settings}
    \centering

\begin{tabularx}{\textwidth}{|
  >{\raggedright\arraybackslash\hsize=\hsize}X | 
  >{\raggedright\arraybackslash\hsize=\hsize}X | 
  >{\raggedright\arraybackslash\hsize=\hsize}X | 
  >{\raggedright\arraybackslash\hsize=\hsize}X |
  >{\raggedright\arraybackslash\hsize=\hsize}X | 
  >{\raggedright\arraybackslash\hsize=\hsize}X |
}
    \hline
    \textbf{Project} & \textbf{Model} & \textbf{Temperature} & \textbf{Input data format} & \textbf{Output data format} & \textbf{Location of data, prompts, code, results} \\
    \hline
    Seedlists & Claude 3 Opus (from Anthropic AI) & 0 & Text extracted from PDF documents, OCR text from scanned documents & JSON & \url{https://github.com/UtrechtUniversity/generative-ai/tree/main/seedlists} \\
    & & & & & \\
    & Assistants API (beta) with gpt-4-0125-preview model (from OpenAI) & Could not be set & Files in PDF and text format uploaded directly to the API  & JSON & Table \ref{tab:seedlist_gpt4_3runs} \\
    \hline
    Health Technology Assessment (HTA) documents & Claude 3 Opus (from Anthropic AI) & 0 & Text extracted from PDF documents & JSON & \url{https://github.com/UtrechtUniversity/generative-ai/tree/main/hta} \\
    \hline
    Kickstarter & Claude 3 Opus (from Anthropic AI) & 0 & Text fields extracted from CSV file & JSON &  \url{https://github.com/UtrechtUniversity/generative-ai/tree/main/kickstarter} \\
    \hline
\end{tabularx}
\end{table*}

\section{Use case: Seedlists}
\label{sec:seedlists}

\subsection*{Scientific goal of the project}

This project aims to digitise the historical seedlist archive of the Utrecht University Botanic Gardens. This archive contains seedlists from botanical gardens all over the world, from the last $\sim$200 years. The goal is to create a database of plant species present at various botanical gardens over the world, over time.  This information can then be further analysed for various scientific purposes, such as determining best practices regarding curational policies and gaining insight into the effects of climate change on plant species. 

\subsection*{Research Engineering task} 

The Research Engineering task is to extract the plant species names from these seedlists. A plant species name consists of the genus, epithet, subspecies name (if present), variety name (if present), form name (if present), cultivar name (if present), author name(s) (if present) and synonym (if present).

\subsection*{Input data} 

The seedlists fall into two categories: 
\begin{enumerate}
\item Newer seedlists (after $\sim$2010) in digital format, usually available as PDF or Excel files.
\item Older seedlists (before $\sim$2010), of which there are only paper versions -- printed, typewritten, or handwritten. These have been scanned, and Optical Character Recognition (OCR) has been performed on them with ABBYY FineReader PDF\footnote{https://pdf.abbyy.com/}, at the Utrecht University library. Thus we have the text from these seedlists, albeit with OCR errors. 
\end{enumerate}

We work with these digitised seedlists, from which a computer can read the text, as input data. 

This project is in the pilot phase, in which there are $\sim$2000 seedlists which give rise to $\sim$45,000 pages to be processed. The entire archive contains $\sim$30,000 seedlists.  

\subsection*{Why use generative AI?}

\begin{figure*}[htbp]
  \centering
  \subfloat[]{\includegraphics[width=0.33\textwidth]{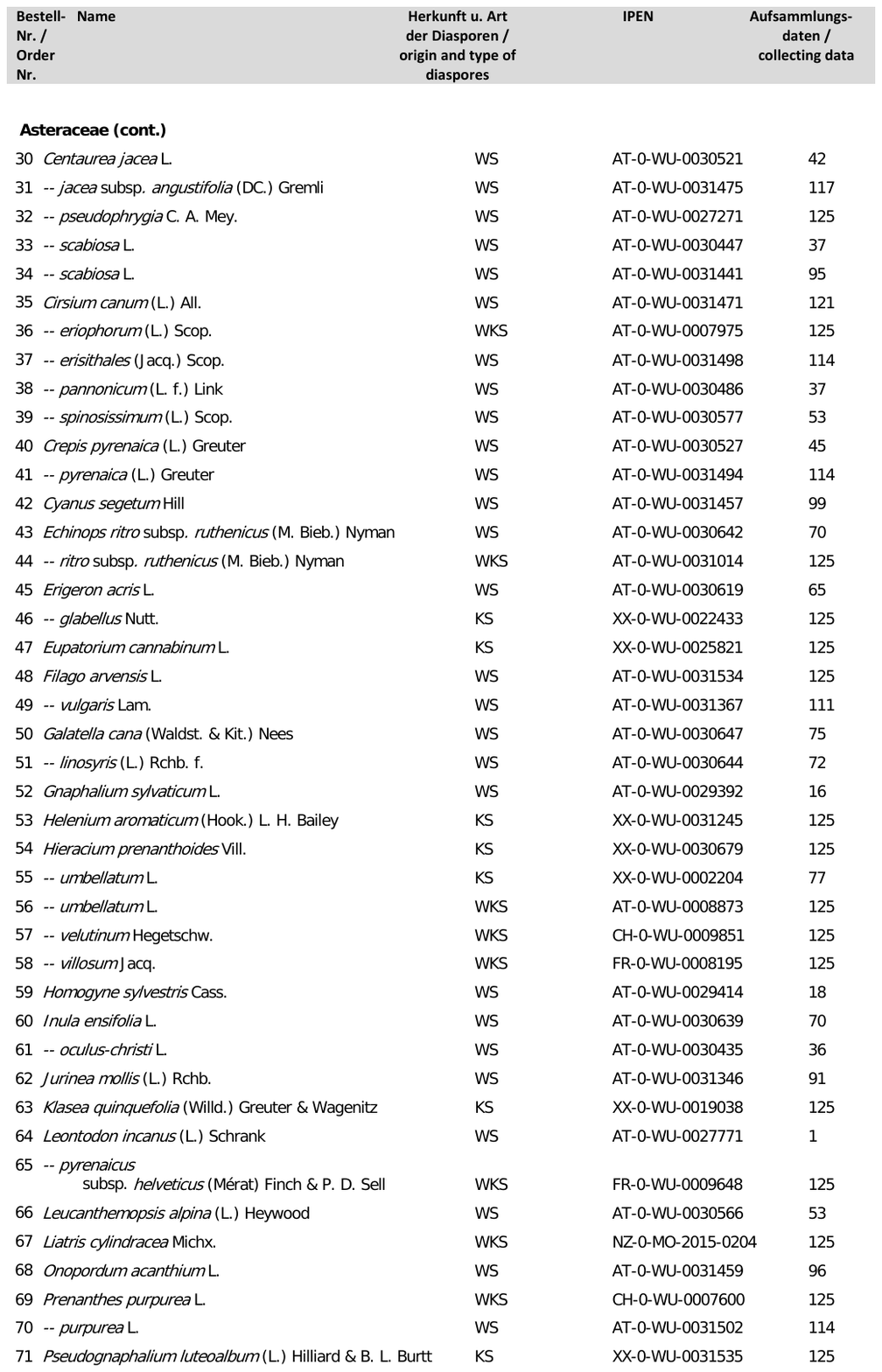}\label{fig:sub12}} 
  \subfloat[]{\includegraphics[width=0.33\textwidth]{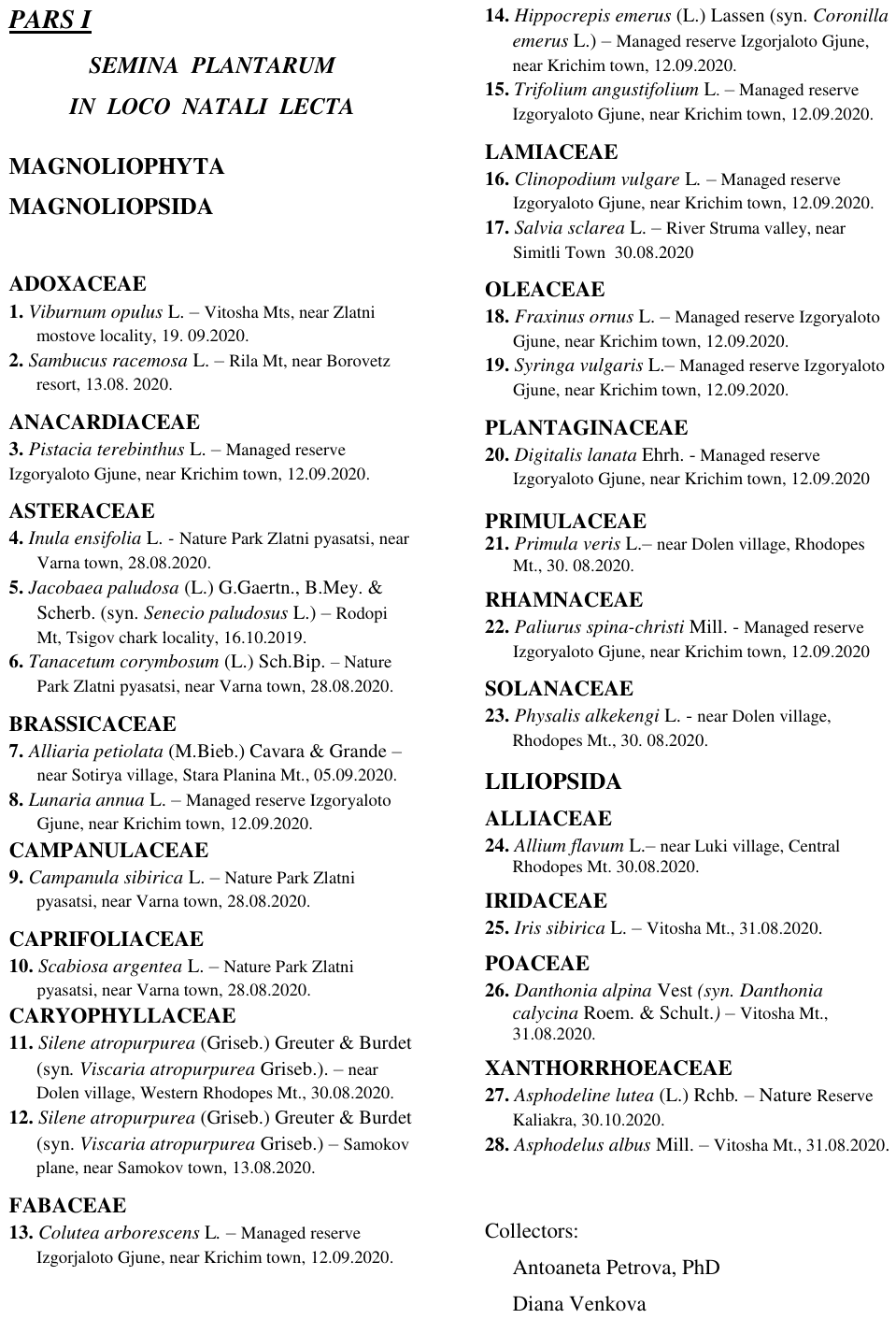}\label{fig:sub2}} 
  \subfloat[]{\includegraphics[width=0.33\textwidth]{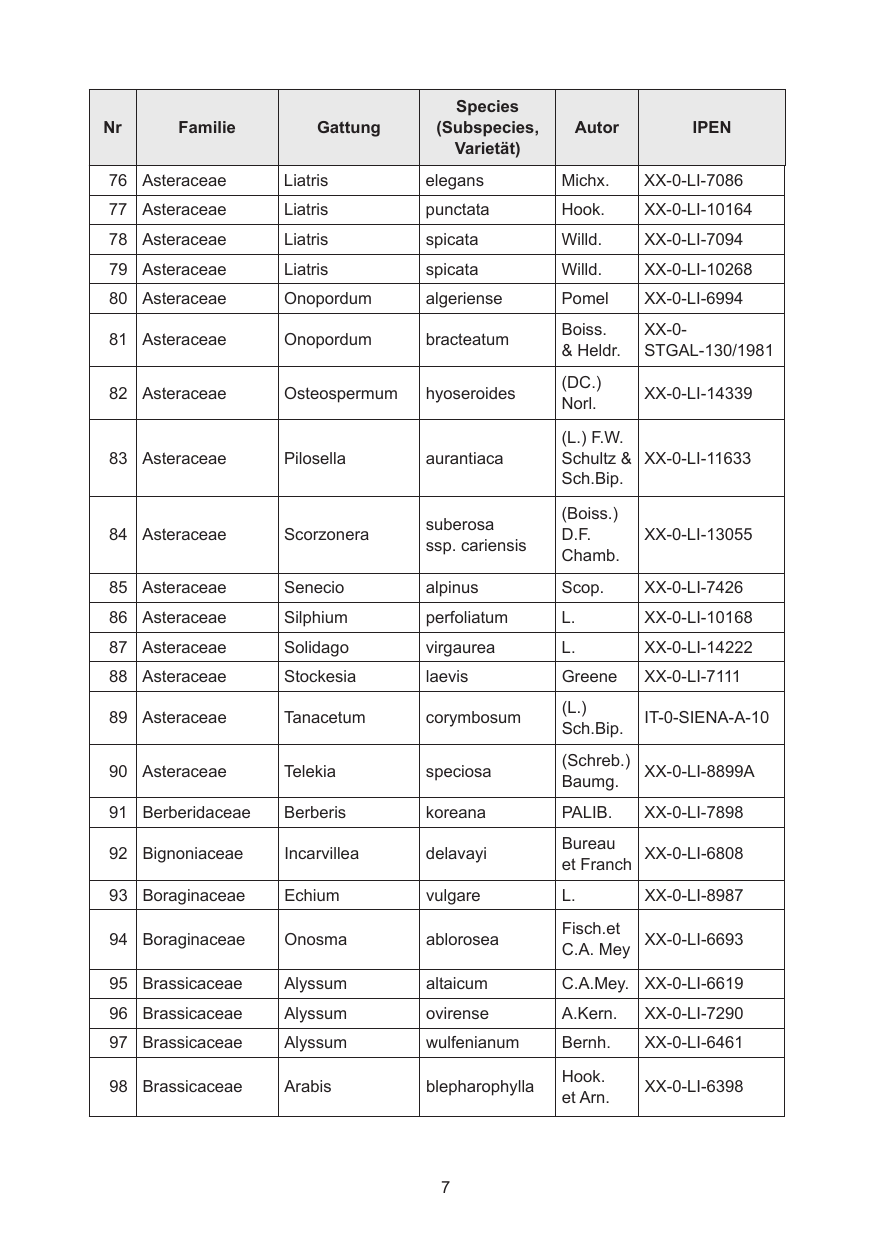}\label{fig:sub9}} 
  \caption{Pages from seedlists in PDF format, from different Botanical Gardens, for the year 2020. These illustrate the diversity of seedlist formats. 
  (\ref{fig:sub12}) Botanischer Garten der Universität Wien, Vienna, Austria
  (\ref{fig:sub2}) Hortus Botanicus, Academiae Scientiarum Bulgariae, Sofia, Bulgaria
  (\ref{fig:sub9}) Botanischer Garten und Arboretum, Linz, Austria
  }
  \label{fig:botanical_gardens_seedlists}
\end{figure*}

\begin{figure}[htbp]
  \centering
  \resizebox{0.48\textwidth}{!}{%
  \subfloat[Scanned text]{\includegraphics[width=0.48\columnwidth]{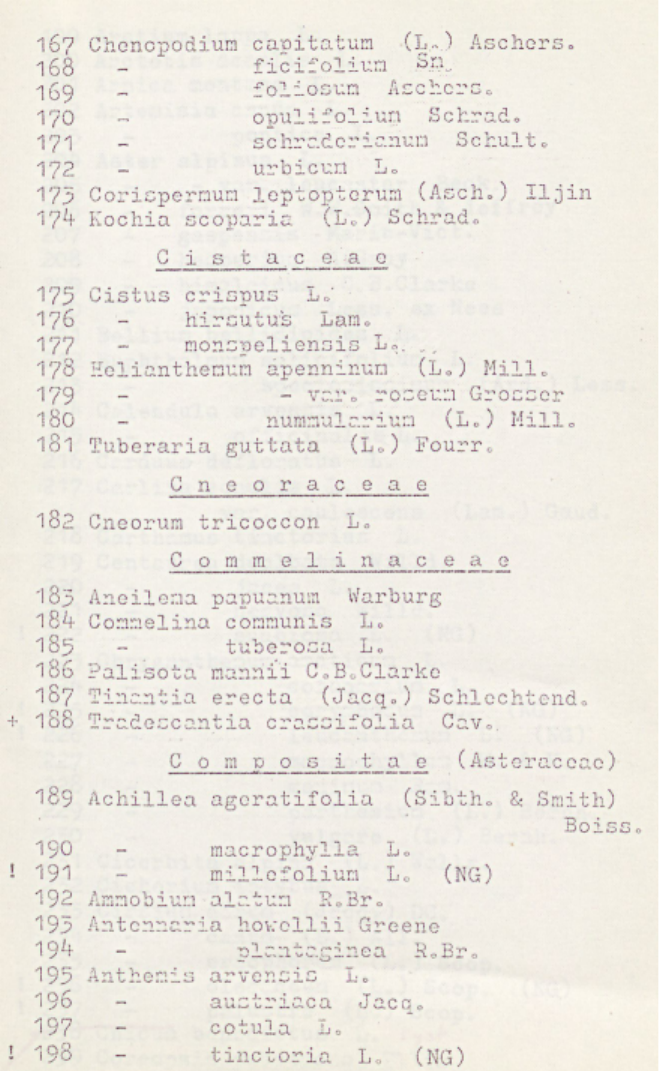}\label{scan}}
  \hspace{0.02\columnwidth}
  \subfloat[OCR text]{\includegraphics[width=0.4\columnwidth,height=0.3\textheight]{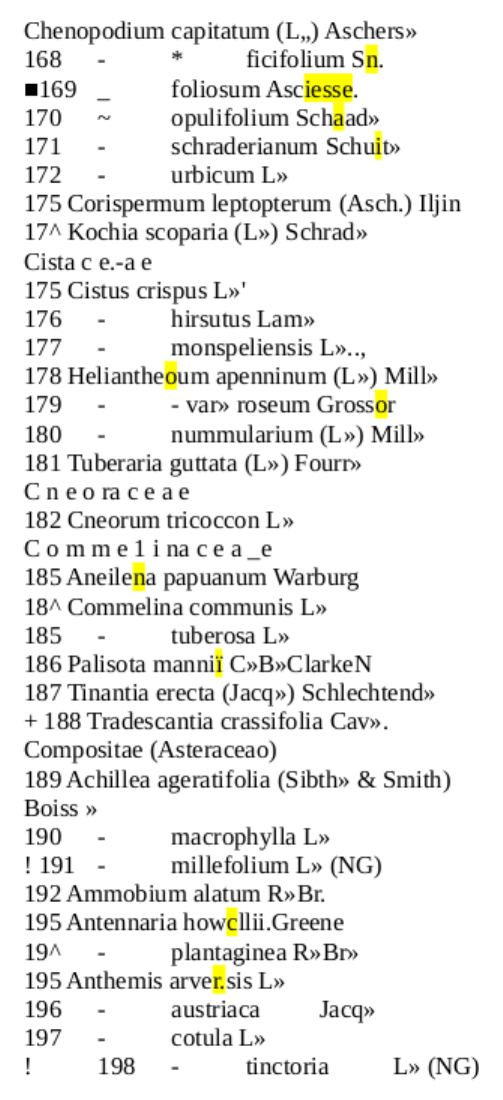}\label{ocr}} 
  }
  \caption{A page from a scanned seedlist from the Botanical Garden of the University of Göttingen, 1970: 
  (\ref{scan}) Original scanned text 
  (\ref{ocr}) Text obtained by performing OCR on the scanned text. The OCR errors are highlighted in yellow.
  }
  \label{fig:botanical_gardens_seedlist_scanned}
\end{figure}

The seedlists are in many different kinds of formats. Some present the botanical information as a one-column list, some as a two-column list, some as a table, etc. This is illustrated in Fig. \ref{fig:botanical_gardens_seedlists}, which shows pages from three newer seedlists. Fig. \ref{fig:botanical_gardens_seedlist_scanned} shows a page from an older typewritten seedlist and the text obtained from it after OCR. While machine-reading this information, it is often difficult to extract the plant species name because it is embedded within other text, and/or there are several epithets following a single genus, as can be seen in Fig. \ref{fig:sub12}. This makes it very difficult to extract the plant species names using a rule-based approach – because there are no rules regarding structure, format, etc. that apply consistently to all the seedlists.  

Therefore, we use generative AI since it is known to be able to interpret unstructured text and extract information from it.

\subsection*{How generative AI is used}

We provided the input data in text format to the Claude 3 Opus model. We converted the seedlists in digital format to text. For the newer seedlists in PDF format, we used the python package PyPDF\footnote{https://pypdf.readthedocs.io/en/stable/} to extract the text. For the older seedlists, the OCR extracted from the scanned seedlists was already in text format. We combined this input data with instructions for data processing to arrive at the final prompt we provided to generative AI. 

\subsection*{Results and evaluation} 
We evaluated the accuracy by constructing a ground truth by interpreting the tables as a human reader and noting all the species present, and then manually comparing this ground truth with the results produced by generative AI. The number of plant species names in Figs. \ref{fig:sub12}, \ref{fig:sub2}, \ref{fig:sub9}, and \ref{scan} is 42, 28, 23, and 32, respectively. Generative AI was able to extract all these species names, and it was able to extract each species name correctly. Thus for this very limited sample of seedlist pages, the recall, precision, and accuracy of our method are all 100\%. We performed this process three times for all these four seedlist pages, and got consistent results. 

For the text obtained from OCR of scanned text (Fig. \ref{ocr}), there are OCR errors, which generative AI was able to correct. We note that the OCR for this document is of good quality -- while there are errors, the character error rate (CER) in Fig. \ref{ocr} is low ($<\sim3\%$). 

While the results from this small sample are promising, this is no guarantee that the results will be accurate for all seedlists or that they will be consistent for all possible runs. However, this level of accuracy and consistency gives us confidence in the value of generative AI as a reliable data processing tool for this task. 

While we settled on Claude 3 Opus as our model of choice because of its superior performance, we also present our earlier attempts with OpenAI's Assistants API\footnote{\url{https://platform.openai.com/docs/assistants/overview}} with one of the models from their GPT-4 series of models, to illustrate the inaccuracies and generative variability that might be present in outputs produced by generative AI. OpenAI's Assistants API is a tool still in beta mode. It allows one to directly upload a file and ask generative AI questions about it. At the time we performed this experiment (March 2024), the temperature could not be specified while using the Assistants API. 
 
We used the Assistants API with the gpt-4-0125-preview model for extracting the species names from the OCR text shown in Fig. \ref{ocr}. The Assistants API was able to extract all the species names, and even correct some OCR errors. However, it also introduced some new errors. We ran the Assistants API thrice, the results of which are shown in Table \ref{tab:seedlist_gpt4_3runs}. We see that the majority of the extracted species names were accurate and consistent over the three runs. The three runs all produced different errors, but most of these involved abbreviation or expansion of the author name, which we consider harmless. In a few cases more serious errors were introduced, like substitution or inclusion of the wrong text, or exclusion of part of the species name.

\begin{table*}[htbp]
    \caption{Extracted species names from three runs with the Open AI Assistants API, using the model gpt-4-0125-preview. The temperature could not be specified in this case. The parts highlighted in yellow show the errors from OCR that remain in the extracted species names, the ones in green show changed text where the change is harmless (abbreviation or expansion of the author name), and the ones in red  show text where the change is undesirable or erroneous (substitution by wrong text, inclusion of parts of the original text that are not part of the species name, exclusion of parts of the original text that are actually part of the species name).}
    \label{tab:seedlist_gpt4_3runs}
    \centering
    \begin{tabularx}{\textwidth}{|X|X|X|}
    \hline
    \textbf{Run 1} & \textbf{Run 2} & \textbf{Run 3} \\
    \hline
    Chenopodium capitatum (L.) \tightboxg{Asch.}			&	Chenopodium capitatum (L.) Aschers.				&	Chenopodium capitatum (L.) \tightboxg{Asch.}				\\ 
Chenopodium ficifolium Sm.				&	Chenopodium ficifolium Sm.				&	Chenopodium ficifolium Sm.				\\
Chenopodium foliosum Asc\tightboxy{iesse.}				&	Chenopodium foliosum Asc\tightboxy{iesse.}				&	Chenopodium foliosum \tightboxg{Asch.}				\\
Chenopodium opulifolium Schrad.				&	Chenopodium opulifolium Schrad.				&	Chenopodium opulifolium Schrad.				\\
Chenopodium schraderianum Sch\tightboxr{arn}t.				&	Chenopodium schraderianum Schult.				&	Chenopodium schraderianum Schult.				\\
Chenopodium urbicum L.				&	Chenopodium urbicum L.				&	Chenopodium urbicum L.				\\
Corispermum leptopterum (Asch.) Iljin				&	Corispermum leptopterum (Asch.) Iljin				&	Corispermum leptopterum (Asch.) Iljin				\\
Kochia scoparia (L.) Schrad.				&	Kochia scoparia (L.) Schrad.				&	Kochia scoparia (L.) Schrad.				\\
Cistus crispus L.				&	Cistus crispus L.				&	Cistus crispus L.				\\
Cistus hirsutus Lam.				&	Cistus hirsutus Lam.				&	Cistus hirsutus Lam.				\\
Cistus monspeliensis L.				&	Cistus monspeliensis L.				&	Cistus monspeliensis L.				\\
Helianthemum apenninum (L.) Mill.				&	Helianthemum apenninum (L.) Mill.				&	Helianthemum apenninum (L.) Mill.				\\
Helianthemum apenninum var. roseum \tightboxg{Gross.}				&	Helianthemum \tightboxr{\phantom{apennium}} var. roseum Grosser				&	Helianthemum apenninum  \tightboxr{(L.) Mill.} var. roseum \tightboxg{Gross.}				\\
Helianthemum nummularium (L.) Mill.				&	Helianthemum nummularium (L.) Mill.				&	Helianthemum nummularium (L.) Mill.				\\
Tuberaria guttata (L.) Fourr.				&	Tuberaria guttata (L.) Fourr.				&	Tuberaria guttata (L.) Fourr.				\\
Cneorum tricoccon L.				&	Cneorum tricoccon L.				&	Cneorum tricoccon L.				\\
Aneilema papuanum \tightboxg{Warb.}				&	Aneile\tightboxy{n}a papuanum Warburg				&	Aneile\tightboxy{n}a papuanum \tightboxg{Warb.}				\\
Commelina communis L.				&	Commelina communis L.				&	Commelina communis L.				\\
Commelina tuberosa L.				&	Commelina tuberosa L.				&	Commelina tuberosa L.				\\
Palisota \tightboxr{barteri Hook.}				&	Palisota mannii C.B.Clarke				&	Palisota mannii C.B.Clarke				\\
Tinantia erecta (Jacq.) \tightboxg{Schlechtendal}				&	Tinantia erecta (Jacq.) Schlechtend.				&	Tinantia erecta (Jacq.) Schlechtend.				\\
Tradescantia crassifolia Cav.				&	Tradescantia crassifolia Cav.				&	Tradescantia crassifolia Cav.				\\
Achillea ageratifolia (Sibth. \& \tightboxg{Sm.}) Boiss.				&	Achillea ageratifolia (Sibth. \& Smith) Boiss.				&	Achillea ageratifolia (Sibth. \& \tightboxg{Sm.}) Boiss.				\\
Achillea macrophylla L.				&	Achillea macrophylla L.				&	Achillea macrophylla L.				\\
Achillea millefolium L.				&	Achillea millefolium L.				&	Achillea millefolium L. \tightboxr{(NG)}				\\
Ammobium alatum R.Br.				&	Ammobium alatum R.Br.				&	Ammobium alatum R.Br.				\\
Antennaria \tightboxr{dioica (L.) Gaertn.}				&	Antennaria howellii Greene				&	Antennaria howellii Greene				\\
Antennaria plantaginea R.Br.				&	Antennaria plantaginea R.Br.				&	Antennaria plantaginea R.Br.				\\
Anthemis arvensis L.				&	Anthemis arvensis L.				&	Anthemis arvensis L.				\\
Anthemis austriaca Jacq.				&	Anthemis austriaca Jacq.				&	Anthemis austriaca Jacq.				\\
Anthemis cotula L.				&	Anthemis cotula L.				&	Anthemis cotula L.				\\
Anthemis tinctoria L.				&	Anthemis tinctoria L.				&	Anthemis tinctoria L. \tightboxr{(NG)}				\\
\hline
\end{tabularx}
\end{table*}

\section{Use case: Health Technology Assessment (HTA) documents}
\label{sec:hta}
\subsection*{Scientific goal of the project}
This project is from the Pharmaceutical Sciences department at Utrecht University. 

 Reimbursement documents published by HTA organisations assess the clinical effectiveness and cost-effectiveness of new drugs, and provide guidance and recommendations for use by policymakers, healthcare providers, and insurance companies. This project aims to extract data points from reimbursement documents published by HTA bodies in the EU. The goal is to create an Open Science database that can be used by other researchers and decision makers. 

This project aims to carry out this task initially for documents from the UK, France, and the Netherlands, and then to extend this to other EU countries. Table \ref{tab:hta_data_points} lists the data points to be extracted from the HTA documents. 

\begin{table*}[htbp]
\centering
\caption{Data points to be extracted from HTA documents}
\label{tab:hta_data_points}
    \centering
\begin{tabularx}{\textwidth}{|
  >{\raggedright\arraybackslash\hsize=.75\hsize}X | 
  >{\raggedright\arraybackslash\hsize=1.25\hsize}X | 
}
\hline
\textbf{Data point} & \textbf{Explanation} \\
\hline
HTA ID & Name of HTA organisation performing the assessment \\
Assessment type & Is this the first assessment, a reassessment, or an indication broadening? \\
Internal identifier & Code or label identifying the document \\
INN & International non-proprietary name of assessed drug \\
Brand name & Brand name of assessed drug \\
Assessment date & When was the assessment finalised? \\
Indication & Medical condition for which the drug is assessed \\
Final recommendation & What is the final recommendation for this drug-indication combination? \\
Comparator & Drug(s) with which the performance of the assessed drug is compared \\
Relative effectiveness assessment outcome & Outcome of the relative effectiveness assessment for this drug-indication combination \\
Cost-effectiveness assessment outcome & Outcome of the cost-effectiveness assessment for this drug-indication combination \\
Budget impact outcome & Budget impact of adoption of the drug \\
Managed entry agreements & Was any OECD-defined managed entry agreement proposed? If so, which class? \\
Clinical restrictions & Clinical restrictions stated in the recommendation \\
\hline
\end{tabularx}
\end{table*}

\subsection*{Research Engineering task} 

The Research Engineering task is to extract the data points listed in Table \ref{tab:hta_data_points}.

\subsection*{Input data} 

The reimbursement documents constitute the input data. These are in PDF format, obtained from the websites of NICE (National Institute for Health and Care Excellence, UK),\footnote{https://www.nice.org.uk/} HAS (Haute Autorité de santé, France),\footnote{https://www.has-sante.fr/} and ZIN (Zorginstituut Nederland, The Netherlands).\footnote{https://www.zorginstituutnederland.nl/}

\subsection*{Why use generative AI?}

The reimbursement documents are in a variety of different formats, having been created by different organisations in different countries, and are in different languages. For some of the data points, the information is not directly or concisely stated in the document. In such cases, one needs the data processing tool to have the ability to comprehend the content and then formulate the answer in a succinct manner. These issues make it difficult to devise a rule-based approach for extracting the required data points from these documents.  

Therefore, we use generative AI since it is known to be able to answer questions about text provided to it \cite{lewis2018generative}. 

\subsection*{How generative AI is used}

We used the PyPDF package to convert the PDF documents to text format, and then used this text as the input to generative AI.

We constructed and used a prompt to extract the desired data points. For the non-English HTA documents, not all the fields of the output JSON object were returned in English. In these cases, we used a second prompt 
to translate the non-English fields.

\subsection*{Results and evaluation}

We present the results for one drug-indication combination, Ivabradine for treating chronic heart failure, assessed by each of the three HTA bodies, NICE \cite{NICE_TA267}, HAS \cite{HAS_Procoralan}, and ZIN \cite{Zorginstituut_Ivabradine}. We constructed a ground truth by interpreting the documents as a human reader and extracting the desired data points. We compared the answers produced by generative AI with this ground truth to determine the accuracy of the results produced by generative AI. We performed three runs with each document to assess the consistency of the results.

Out of the 14 data points to be extracted, 11 were extracted accurately and consistently over all three documents and all three runs. For one document (ZIN), two data points (final recommendation, budget impact outcome) were not lexically identical over the three runs but were still semantically consistent. For one document (HAS), one data point (HTA ID) was consistent but not the answer we desired -- the committee that performed the evaluation,  Commission de la Transparence, was reported, rather than the parent agency, HAS. This stems from the ambiguity in the question ``Which HTA body is performing the assessment?" posed to generative AI.

\section{Use case: Kickstarter}
\label{sec:kickstarter}

\subsection*{Scientific goal of the project}

This project is from the Human Geography and Spatial Planning department at Utrecht University. The goal of this project is to assess whether crowdfunding contributes to local economic growth in the United States, and thus fosters innovation and economic development. 

\subsection*{Research Engineering task}

Kickstarter\footnote{\url{https://www.kickstarter.com/}} is a crowdfunding website for raising money for creative projects. The North American Industry Classification System (NAICS\footnote{\url{https://www.census.gov/naics/}}) is a standard for classifying businesses based on their type of economic activity. The Research Engineering task is to assign a NAICS code (from the year 2017) to every Kickstarter project from 2014 to 2023. This will identify the industry sector, which will facilitate further analysis to determine the economic growth per sector, per year, per county. 

\subsection*{Input data}

Web scraped data of Kickstarter projects are available online \cite{webrobots_kickstarter}. The following four fields in these data are relevant for our task of assigning NAICS codes: 
\begin{enumerate}
\item Name (Title of the project) 
\item Blurb (A brief description of the project) 
\item Kickstarter category 
\item Kickstarter subcategory 
\end{enumerate}
The total number of projects is $\sim$300,000. 

\subsection*{Why use generative AI?}

 Assignment of a NAICS code to a business description is a difficult task. One needs an industry expert with knowledge and understanding of NAICS codes to carry out this task. There are 311 4-digit NAICS codes, so this is a classification problem with a huge number of classes. With the vast number of Kickstarter projects we want to label, this task is impossible to carry out manually. Therefore, we use generative AI for this task.
 
\subsection*{How generative AI is used}

We provided the input data for each project (name, blurb, category, subcategory) to generative AI, and asked it to return the NAICS code it determined to be the most appropriate for the project, based on these four characteristics.

\subsection*{Results and evaluation}

For this project, the results are difficult to evaluate, because there is no ground truth with which the results can be compared. Assessment of the NAICS code is inherently subjective; different human raters might assign different NAICS codes to the same project. This ambiguity makes it difficult to evaluate the performance of generative AI.

We used the concept of interrater reliability \cite{interrater_reliability} to assess the performance of generative AI. A sample of 540 representative projects was chosen, with roughly equal number of projects from each of the 15 Kickstarter categories. These projects were divided into six partially overlapping subsets and assigned to six human raters in a staggered manner, in such a way that each project was assigned a NAICS code by two independent human raters. The NAICS code produced by generative AI was then compared to those produced by the human raters. 

The highest fraction of NAICS codes that matched between generative AI and a (single) human rater was 53\%, over 145 projects. The highest fraction of codes that matched between two human raters was 60\%, over 63 projects. Therefore, for this task, the performance of generative AI is broadly comparable to that of a human rater.

Table \ref{table:ks_nonmatching} shows a sample of projects for which the NAICS codes produced by generative AI differed from those produced by a human rater. It can be seen that both sets of codes are plausible, illustrating the ambiguity inherent in this classification task, and the resulting difficulty of evaluating the quality of the results produced by generative AI -- or by any other method, in fact.

\begin{table*}[htpb]
\centering
\caption{Kickstarter projects for which generative AI and a human rater assigned different NAICS codes, but both codes can be considered appropriate}
\label{table:ks_nonmatching}
\begin{tabularx}{\textwidth}{|
  >{\raggedright\arraybackslash}X | 
  >{\raggedright\arraybackslash}X | 
  >{\raggedright\arraybackslash\hsize=.5\hsize}X | 
  >{\raggedright\arraybackslash\hsize=.5\hsize}X | 
  >{\raggedright\arraybackslash}X | 
  >{\raggedright\arraybackslash}X |
}
\hline
\textbf{Name} & \textbf{Blurb}  & \textbf{Category} & \textbf{Subcategory} & \multicolumn{2}{c|}{\textbf{NAICS code (2017) and category description}} \\
\cline{5-6}
&&&& Human rater & Claude 3 Opus \\
\hline
Inspired &	Inspired by a lifetime of collaboration, I am finally recording an album of my own with the help of great musicians \& friends. &	Music &	Jazz &	7111: Performing Arts Companies &	5122: Sound Recording Industries \\
Let's write poetry together. &	A poetry book from your ideas, written by me. &	Publishing &	Poetry &	7115: Independent Artists, Writers, and Performers	& 5111: Newspaper, Periodical, Book, and Directory Publishers \\
MoonRay - World's Best Desktop 3D Printer &	Meet the 3D Printer you have been waiting for, one that doesn't compromise on anything. & Technology & 3d printing	& 3344: Semiconductor and Other Electronic Component Manufacturing &	3332: Industrial Machinery Manufacturing \\
Cult Party: Support Your Local Feminist Girl Gang &	All Woman Intersectional Feminist Collective. & Fashion & Accessories	& 3152:	Cut and Sew Apparel Manufacturing &	4483: Jewelry, Luggage, and Leather Goods Stores \\
EPIC BITES CURRY KETCHUP &	Epic Bites Curry Ketchup is tradition, cult, and taste and new jobs. With this project, we want to create new jobs.	& Food &	Food trucks & 	3114: Fruit and Vegetable Preserving and Specialty Food Manufacturing &	7225: Restaurants and Other Eating Places \\
\hline
\end{tabularx}
\end{table*}

\section{Discussion and Conclusion}
\label{sec:discussion}

\subsection{Is generative AI an appropriate tool?}
We applied generative AI to process textual data in three use cases, involving complex information extraction, natural language understanding, and text classification tasks. The results are not always completely accurate or consistent, but the amount of serious errors is limited. We observed that generative AI is a good tool for tasks that are easy for a human to perform, but difficult for a computer program to carry out. This is often the case with heterogeneous, unstructured data, for which it is difficult to come up with rules which apply to all data samples -- for example, the different formats for the seedlists, or the different formats and languages for the HTA documents. 

In conclusion, generative AI can be considered as a possible tool for a data processing task if the amount of data to be processed is large, no simple, rule-based method for performing the data processing can be found, and the results are of sufficiently high quality for the research purpose.\footnote{We should stress that we have not attempted to assess the appropriateness of using generative AI in the light of legal and ethical concerns such as those listed in Table \ref{tab:lit-concerns}. We have only assessed the technical appropriateness.}

\subsection{Factors affecting results}
While accuracy and consistency are independent metrics determining the reliability of the results, both are affected by the following factors:
\begin{enumerate}

\item \textit{Temperature:} 
The lower the value of the temperature parameter, the higher the accuracy and consistency of the outputs. For research data processing tasks for which the desired results are well-defined and objective, a temperature of 0 maximises the accuracy and consistency of the outputs.

\item \textit{Prompt:} 
A  well-posed, unambiguous, and clear prompt increases the chance of getting accurate and consistent results. 
One needs to carefully craft the prompt so that one gets the desired results. %One should try to pose closed-ended questions or questions with very well-defined answers, if possible. 
This part, known as \textit{prompt engineering} \cite{anthropic_prompt_engineering, OpenAI_prompt_engineering}, is typically the most time-consuming part of setting up a generative AI pipeline. One needs to iterate over many different versions of the prompt, with a representative input dataset, to arrive at the optimal prompt that will work in a general way on the entire dataset. 

\item \textit{Input data:} 
The more obviously a fact is stated in the input document, the better generative AI can extract it. If the answer to a question requires ``reading between the lines", analysing or synthesising several parts of a document to arrive at a conclusion, or very specific domain knowledge in a field not well-represented in its training data, generative AI is likelier to provide a poorer-quality answer.
\end{enumerate}

\subsection{Future work}
This study, in which we demonstrate the application of generative AI via illustrative examples, is largely qualitative in nature. In the future we plan to perform a more quantitative evaluation of the performance of generative AI in research data processing tasks. We plan to apply generative AI to process a larger, statistically significant number of data samples from the three use cases, and calculate standard metrics like precision, recall, and accuracy. This will provide a quantitative assessment of the performance of this method.  

\section*{Acknowledgment}

The seedlists project was provided by Martin Smit and Edwin Pos, the HTA project by Jan-Willem Versteeg and Lourens Bloem, and the Kickstarter project by Nicola Cortinovis.

The paper seedlists were scanned by Christi Wagner and Martin Smit, and the OCR was performed by Coen van der Stappen.

We thank Roel Brouwer, Maarten Schermer, and Ingo Schröder for carefully reading the manuscript and providing valuable feedback which improved this paper.

\bibliographystyle{IEEEtran}
\bibliography{GenAI}

\end{document}